\documentclass[final]{cvpr}
\usepackage{times}
\usepackage{epsfig}
\usepackage{graphicx}
\usepackage{amsmath}
\usepackage{amssymb}
\usepackage{float}
\usepackage{multirow}
\usepackage[table,x11names]{xcolor}
\usepackage{array}

\usepackage[pagebackref=true,breaklinks=true,colorlinks,bookmarks=false]{hyperref}

\def\cvprPaperID{7368} 
\def\confYear{CVPR 2021}

\begin{document}
    \setlength{\abovedisplayskip}{3pt}
    \setlength{\belowdisplayskip}{3pt}
	
	\title{Self-Supervised Simultaneous Multi-Step Prediction of Road \\ Dynamics and Cost Map} 
\author{Elmira Amirloo$^1$, Mohsen Rohani\thanks{The authors contributed equally.} ${}{}{  }^1$ , Ershad Banijamali$^1$, Jun Luo$^1$, Pascal Poupart$^2$\\
$^1$Noah's Ark Lab, Huawei, Toronto, Canada \\
$^2$School of Computer Science, University of Waterloo, Waterloo, Canada
 \\
{\tt\small  $\{$elmira.amirloo,mohsen.rohani,ershad.banijamali1,jun.luo1$\}$@huawei.com}, 
{\tt\small ppoupart@uwaterloo.ca}
\and

}
	
	\maketitle
	
    \vspace{-30pt}
	\vspace{-20pt}
\begin{abstract}

While supervised learning is widely used for perception modules in conventional autonomous driving solutions, scalability is hindered by the huge amount of data labeling needed. In contrast, while end-to-end architectures do not require labeled data and are potentially more scalable, interpretability is sacrificed. We introduce a novel architecture that is trained in a fully self-supervised fashion for simultaneous multi-step prediction of space-time cost map and road dynamics. Our solution replaces the manually designed cost function for motion planning with a learned high dimensional cost map that is naturally interpretable and allows diverse contextual information to be integrated without manual data labeling. Experiments on real world driving data show that our solution leads to lower number of collisions and road violations in long planning horizons in comparison to baselines, demonstrating the feasibility of fully self-supervised prediction without sacrificing scalability.


\end{abstract}

	\section{Introduction}

Conventional autonomous driving (AD) stacks consist of various modules \cite{urmson2008boss}. A perception component is responsible for detecting objects in the scene and a prediction module for projecting their positions in the future. Based on their outputs a motion planner generates a desired trajectory according to a manually specified cost function \cite{choset2005principles}, which is in turn executed by a controller. A key advantage of this approach is the interpretability of the final decision. For example, in case of accident, each component can be investigated individually. However, with different parts designed and tuned separately, each module is not aware of the errors made by the other parts. In many cases, there is no clear way for estimating the model uncertainty and propagating it to the system. In addition to massive amount of human-labelled data required to train the perception components, the manual design and tuning of the cost function for motion planning tends to limit the system's ability for dealing with complex driving scenarios .

As an alternative, several works \cite{bojarski2016end,codevilla2018end,xu2017end,chi2017deep} proposed driving systems that use raw sensory input to directly produce control commands (i.e., acceleration and steering). This approach allows full backpropagation and eliminates the need for a cost function. Since a large quantity of data can be collected from cars equipped with appropriate sensors and directly used for training without human labelling, this approach is potentially highly scalable with data and compute. However, such a monolithic approach lacks internally interpretable components, offers little insight as to how system faults may arise, and is thus ill-suited for safety-critical real-world deployment.

 \begin{figure}[!t]
	\vspace{-.35cm}
	\centering
	\includegraphics[scale=0.5]{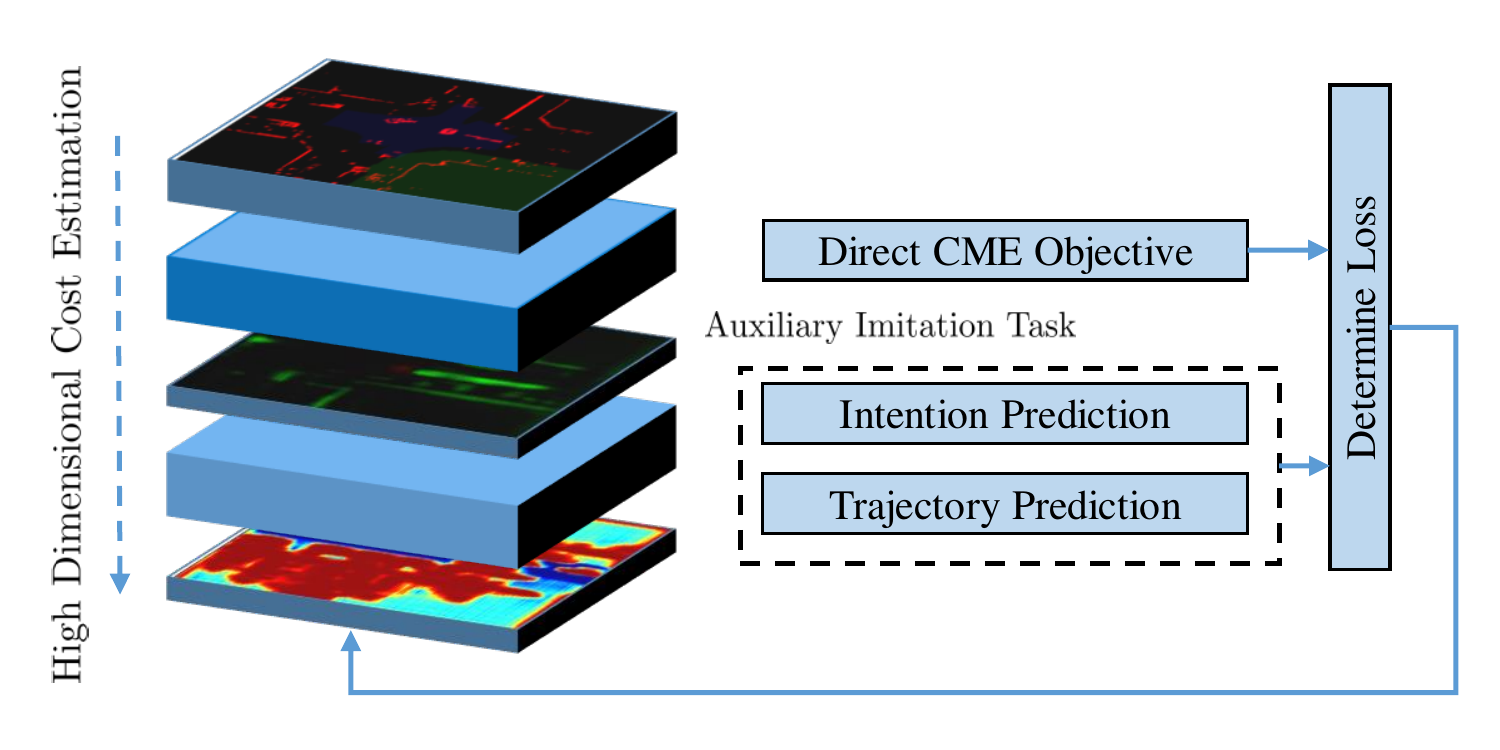}
	\vspace{-.3cm}
	\caption{\small {High Dimensional Cost Map Estimation.}
	}
	\vspace{-.6cm}
	\label{fig:arch}
\end{figure}

In this paper, we propose a new approach that allows meaningful interpretation and avoids manual data labeling and design of cost function. Our approach is centered around a novel architecture for learning-based space-time Cost Map Estimation (CME). 
The proposed method is highly scalable as it can be trained in a fully self-supervised fashion. Moreover, the space-time cost map has both a natural interface with motion planner and an interpretable  domain-specific semantics.

Specifically, our architecture encodes high dimensional Occupancy Grid Maps (OGM)s as well as other contextual information (e.g., drivable area and intersections) and  simultaneously predicts the OGMs and estimate the cost map for multiple steps into the future. This leads to interpretable intermediate representations in the form of OGMs.  A cost map (CM) is a grid map where the value of each cell represents the cost of driving into that cell location. By extending the predicted CM multiple steps into the future, we arrive at a sequence of space-time CMs. These CMs can then be used by a motion planner to rank possible future trajectories through integrating the cost over the cells these trajectories occupy.

Importantly, while it is obvious that OGM prediction training can be made self-supervised using sequences of driving data (e.g., \cite{mohajerin2019multi}) so long as occupancy estimation is accurate, labels for self-supervised training of CM estimation are much harder to synthesize. To solve this problem, we decompose the CME objective into two parts. The first one injects the prior knowledge about the environment where it is available (e.g., occupied cells are high cost). However, there is no explicit information about the cost of most of the cells. Hence, we propose using an auxiliary task to guide the training. Using auxiliary objectives for improving the performance of a model in a primary task has been proven to be effective in different fields \cite{liu2015deep,trinh2018learning,burda2018large}. Similarly, in this work we define an auxiliary imitation task that forces the model to predict the expert's intention and trajectory based on the estimated CMs. For this task a data-driven set of intentions capturing different modes of driving is used. This objective term pushes the model to fill in the blanks and arrive at complete and systematically accurate predictions of the CM.

The main contributions of this paper are as follows:
\begin{itemize}
    \vspace{-.10cm}
    \item An architecture for estimating CMs simultaneously with OGM prediction from human driving data that is fully self-supervised and requires no extra data labeling,
    \vspace{-.10cm}
    \item A set of specific training objectives that combines environment constraints, expert's behavior, map information as well as solving auxiliary imitation tasks leading to estimating space-time cost maps,
    \vspace{-.10cm}
    \item An empirical demonstration of the effectiveness of this design in the overall performance of an AD system through multiple experiments and generalization tests.
\end{itemize}

	\section{Related Work}

Unlike the monolithic end-to-end approach, our proposal replaces modules of a conventional AD stack to enable doing a fully self-supervised learning of cost maps. Our main goal is balancing scalability and interpretability. From this perspective, we discuss how our work compares and contrasts with existing works.

\subsection{Cost Design and State Prediction}

Conventional motion planners typically optimize trajectories according to a predefined cost function \cite{werling2010optimal,schlechtriemen2016wiggling,hu2018dynamic}. However, manually defining and tuning a cost is extremely hard for highly dynamic environments such as driving. This has led to trade-offs where AD solutions largely avoid difficult interactions. More recently learning based received significant attention in planning \cite{paden2016survey}. Similarly, we aim at replacing the manual cost function with learning-based space-time CMs that predict into the future. However, it is important to note that our proposal is compatible with both classical and learning-based planning methods insofar as they can use the predicted CMs to rank candidate trajectories.

Prediction is a crucial part of any AD stack. Learning-based prediction is increasingly popular in AD research. In \cite{luo2018fast}, detection, tracking and motion forecasting are done using a single network. \cite{sadeghian2018car} uses a bird's eye view image of navigation and the motion history of agent to predict its future path. \cite{tang2019multiple, rhinehart2019precog, rhinehart2018r2p2} combine high dimensional and low dimensional data to predict multi-modal trajectories. Both video prediction \cite{xu2017end} and OGM prediction \cite{hoermann2018dynamic, mohajerin2019multi} have been done in AD research. In contrast to such work, our approach aims to simultaneously predict states and estimate CMs. In many cases, there is no clear way to compute the uncertainty of the isolated prediction models and propagate it to the motion planner. Furthermore, although we evaluate the quality of estimated CMs with a specific trajectory sampler, in our design any set of trajectories can be ranked using the estimated CMs. Predicting trajectories directly makes both generalization and adaption to a new or temporary constraint harder.

\subsection{RL and IRL}

As a powerful framework for sequential decision making, reinforcement learning (RL) makes an attractive choice for AD research. \cite{paxton2017combining} proposed a hierarchical RL scheme that divides tasks into high-level decisions and low-level control. \cite{shalev2016long} introduced a two-phase approach for cruising and merging tasks in autonomous driving, where future states are first predicted from the current ones and then an RL planner uses these predictions to output acceleration. These proposals utilize low-dimensional states derived from perceptual processing of high dimensional sensory data. However, by implicitly conflating the processed low dimensional state estimations as the actual states, error and uncertainty do not get propagated through to improve the whole system. Moreover, similar to the case of manual design of the cost function, defining reward for RL solutions remains an open research challenge.

Inverse reinforcement learning (IRL) focuses on learning a reward function from the expert's behavior, with maximum entropy \cite{ziebart2008maximum} being a popular method. However, standard IRL algorithms can not deal with high dimensional data and continuous space. Some works such as \cite{rosbach2019driving} extended these algorithms to use high dimensional data. But linear estimation of reward function can adversely limit generalization of the system. Similar to our work, \cite{wulfmeier2016watch} estimates a high dimensional cost map using a maximum entropy deep IRL framework. There are two main differences between this work and ours. First, the system is not modular which challenges the interpretability in case of failure. Second, in such IRL frameworks training the policy and estimating the reward is done together. Empirically training the new policy with the learned reward does not lead to similar performance. Hence, if the set of actions or the policy model needs to be changed, the learned reward may not lead to similar performance. 

\subsection{Imitation Learning}

Imitation learning (IL) is a standard approach to learning from expert demonstration. It has the advantage of not requiring costly online exploration typically necessary for RL methods. \cite{bojarski2016end,codevilla2018end} follow a monolithic, end-to-end approach where the network receives images and generates control commands. While this approach avoids costly data labeling, direct behavior cloning suffers from cascading errors when dealing with out-of-distribution inputs. To alleviate this issue, \cite{kuefler2017imitating} adopted Generative Adversarial Imitation Learning (GAIL) \cite{ho2016generative}. But the interpretability challenges remain.

Since highly mature control solutions are widely used in AD systems, learning to do level control is often unnecessary and learning based trajectory planners are more practical. \cite{bansal2018chauffeurnet} uses a modular network to generate driving trajectories. They propose novel data augmentation approaches to generate scenarios such as collisions which most real-world datasets lack and give the network a better opportunity to learn to handle such situations. \cite{zeng2019end} also has a modular design where the perception module does 3D object detection and motion forecasting. A cost volume generation is done by another component of the overall network. They use the cost volume to choose a trajectory with the lowest cost. However, in contrast to our proposal, both of these works heavily rely on labeled data for the perception components.
\noindent
	\section{Technical Approach}

We address the problem of predicting a high dimensional cost map by proposing a modular architecture that can be trained end-to-end in a self-supervised fashion. We then use this prediction to evaluate and score different trajectories. The model takes a sequence of LiDAR point clouds and other contextual information (e.g. map) as input. It predicts the future OGMs representing road dynamics and estimate the space-time cost of driving in each cell of the OGM simultaneously. The proposed architecture has two  components: (1) an OGM predictor and (2) a CM estimator. Note that this design adds interpretability because the predicted OGMs over the planning horizon are independently semantically meaningful.

\subsection{Input Encoding and Network Architecture}

On the input side, frames of LiDAR point clouds from the immediate past are first converted into a sequence of OGMs. These OGMs are then transformed to a reference frame attached to the vehicle's current position. Both prediction and cost estimation are done with respect to this coordinate system to avoid unnecessary complexity due to motion of the ego vehicle. Moreover, similar to \cite{bansal2018chauffeurnet}, we encoded semantic information from the HD map such as drivable area and intersection structure in separate channels. We then concatenated the map encodings with OGMs to form the input to the network.

Predicted OGMs are represented by a binary random variable $o_k(i, j) \in \{0, 1\}$ where $o_k(i, j)$ is the occupancy state of the cell at the $i^{th}$ row and the $j^{th}$ column at the $k^{th}$ time step, with 1 for occupied and 0 for empty. $p_o$ is the probability of occupancy associated with each cell. Cost value at each cell $c_k(i, j)$ is coded similarly where 1 and 0 are assigned to high and low cost cells respectively.

We study two alternative architectures. For the \textit{Recurrent Cost Map Estimator} (RCME), we assume the CMs in each time step are conditionally dependent on the information in previous time steps. For the \textit{Multi-Step Cost Map Estimator} (MSCME), we omit this assumption. In Section \ref{sec:exp} we show that these two architectures have similar performance for shorter prediction horizons. The recurrent architecture performs better for longer planning horizons.

\vspace{5pt}
\begin{figure}[h]
	\raggedleft
	\includegraphics[scale=0.36]{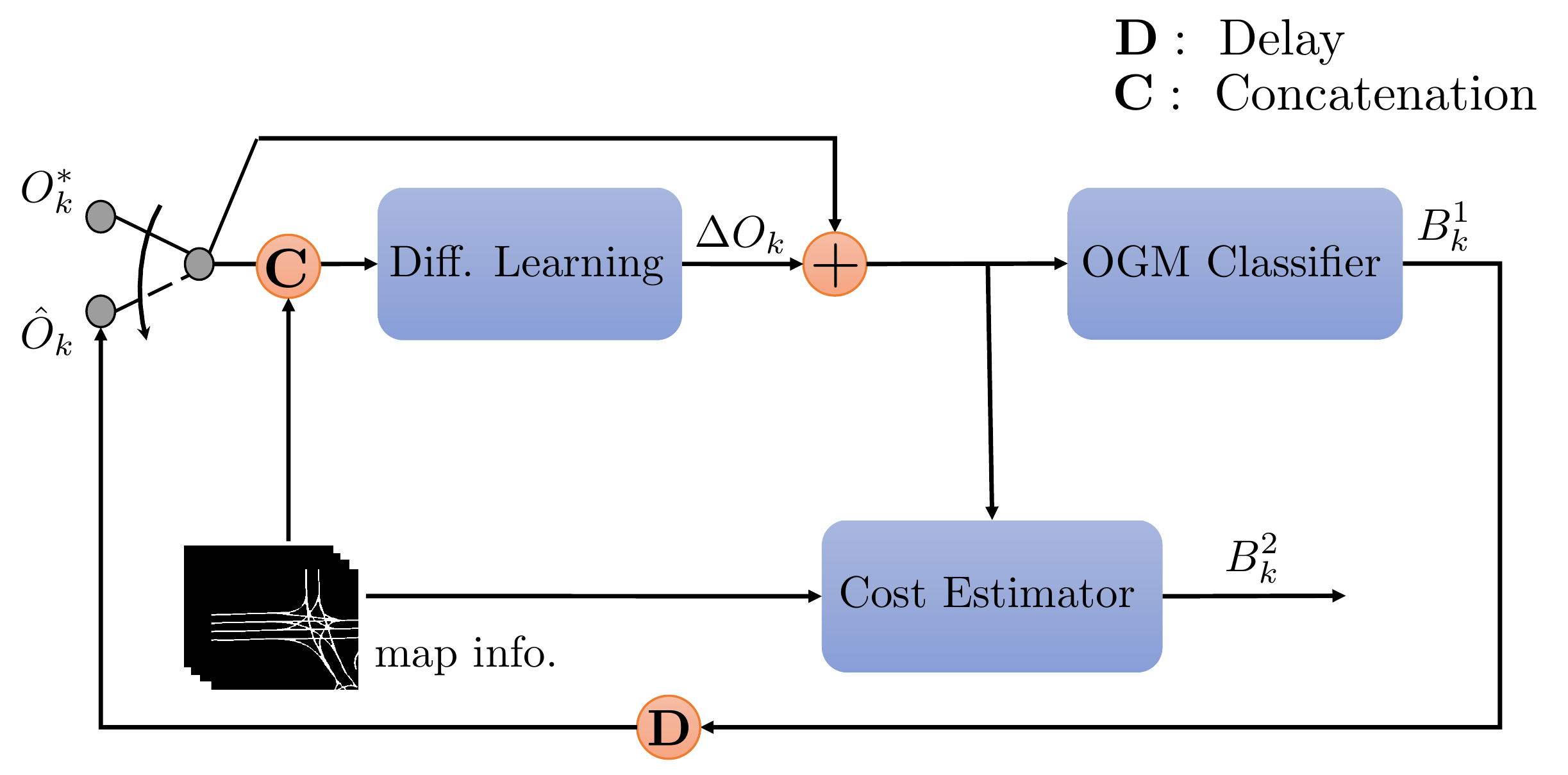}
	\caption{Recurrent Cost Map Estimator}
	\label{fig:rnncost}
\end{figure}

\subsubsection{Recurrent Cost Map Estimator}

The RCME incorporates the \textit{difference learning} method in \cite{mohajerin2019multi} for OGM prediction and extends it with our cost estimator module to simultaneously predict CMs (Figure \ref{fig:rnncost}). Formally, the output of the network at time step $k$ can be represented as a two-channel tensor:

\vspace{1pt}
\begin{equation}
\resizebox{0.9\hsize}{!}{$
\mathbf{B}^1_{k}=\Big{[}p_o\big{(}o_k(i,j)\big{)} \Big{]}=
\begin{bmatrix}
p_o\big{(}o_k(1,1)\big{)} & \dots & p_o\big{(}o_k(1,W)\big{)} \\
\vdots & \ddots & \vdots \\
p_o\big{(}o_k(H,1)\big{)} & \dots & p_o\big{(}o_k(H,W)\big{)} \\
\end{bmatrix}
$}
\label{eq:ogm0}
\end{equation}

\vspace{5pt}
\begin{equation}
\resizebox{0.9\hsize}{!}{$
\mathbf{B}^2_{k}=\Big{[}p_c\big{(}c_k(i,j)\big{)} \Big{]}=
\begin{bmatrix}
p_c\big{(}c_k(1,1)\big{)} & \dots & p_c\big{(}c_k(1,W)\big{)} \\
\vdots & \ddots & \vdots \\
p_c\big{(}c_k(H,1)\big{)} & \dots & p_c\big{(}c_k(H,W)\big{)} \\
\end{bmatrix}
$}
\label{eq:ogm1}
\end{equation}

\noindent
where $B^1_{k}$ and $B^2_{k}$ are the first and second output channels respectively. $o_k(i, j)$ is taken to be independent of the values of other cells at time step $k$, but conditioned on values of all cells in previous time steps. And the same assumption is made for $c_k(i,j)$:
\vspace{-1pt}
\begin{equation}
p_o\big{(} o_k(i,j)|\mathcal{O}_{k-1},\mathcal{O}_{k-2},...\big{)}
\label{eq:probs_1}
\end{equation}
\noindent
\begin{equation}
p_c\big{(} c_k(i,j)|\mathcal{C}_{k-1},\mathcal{C}_{k-2},...\big{)}
\label{eq:probs_2}
\end{equation}
\noindent
where
\noindent
\begin{equation}
\mathcal{O}_k=\big{\lbrace}o_k(m,n)|m=1,...,H;n=1,...,W\big{\rbrace}
\end{equation}
\begin{equation}
\mathcal{C}_k=\big{\lbrace}c_k(m,n)|m=1,...,H;n=1,...,W\big{\rbrace}
\end{equation}

\noindent
and m and n are indices ranging over the entire OGM. The conditional probabilities in Equation \ref{eq:probs_1} and \ref{eq:probs_2} may be captured using the recurrent architecture proposed. In practice, a short history suffices. The network observes OGMs for the past $\tau$ time steps and predicts the OGMs for the next $T$ time steps while estimating a CM at every step.

In reality not every cell in the OGM changes between two time steps. The \textbf{Difference Learning} module implicitly distinguishes between dynamic and static objects. By adding the features extracted by this module to the previous observed OGMs $B^{1*}_{k}$ or predicted OGMs $\hat{B^1_{k}}$ and stacking them with the encoded map, the \textbf{OGM Classifier} can be trained to effectively and efficiently predict if a cell is occupied or not. We did not use the same architecture for estimating CMs as they are not directly \textit{observable} and imposing such a feedback loop can amplify error in CM estimation. Hence, the stacked OGM features are separately fed along with the encoded map to the \textbf{Cost Estimator} module that consists of an encoder and a decoder. The encoder has \{32, 64\} $3\times3$ convolution filters with stride 2. The decoder has two deconvolution layers with \{64, 32\} $3\times3$ filters with stride 2 each deconvolution layer is followed by a convolution layer with the same size and stride 1. 

\subsubsection{Multi-Step Cost Map Estimator}

\begin{figure}[h]
	\raggedleft
	 \includegraphics[scale=0.40]{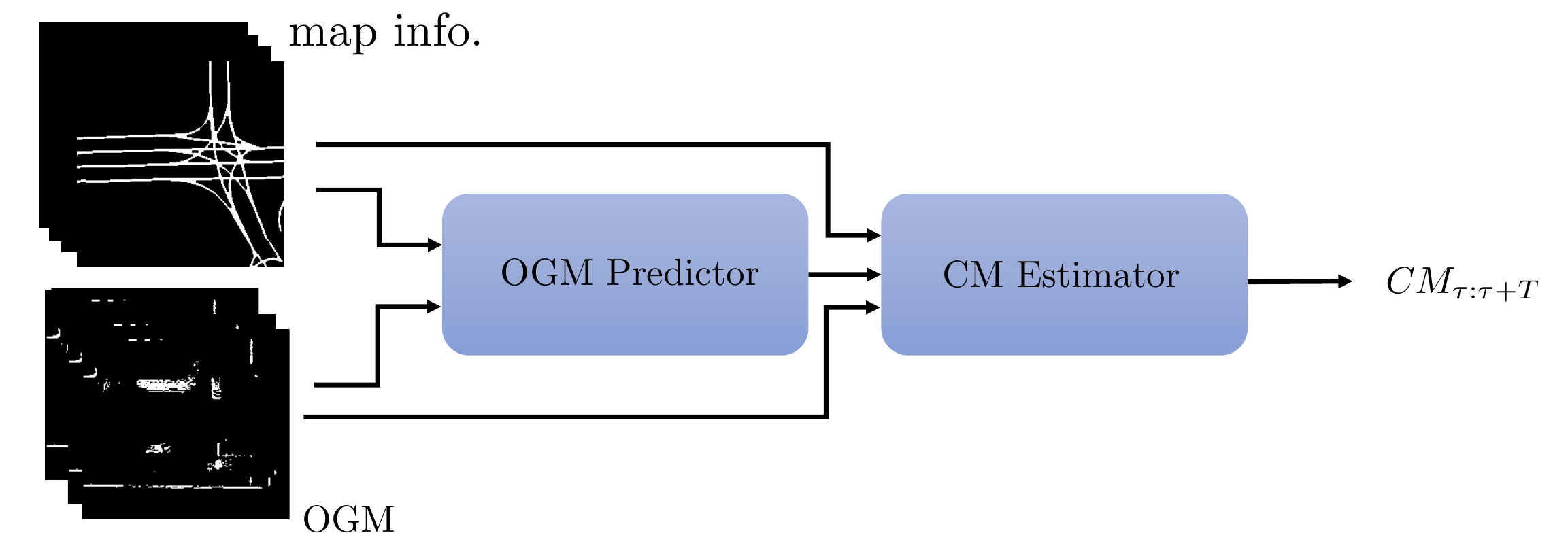}
	 \caption{Multi-Step Cost Map Estimator}
	 \label{fig:mscost}
\end{figure}

The MSCME architecture is illustrated in Figure \ref{fig:mscost}. Similar to RCME, the OGMs and the encoded maps are fed to the predictor network. The predicted and observed OGMs are then stacked with the encoded map and passed through an encoder and then a decoder to estimate a CM for $T$ time steps. To avoid computationally expensive 3D convolutions we concatenate time steps along the last dimension. In order to get similar performance as the previous architecture we used \{32, 64, 128\} filters in the encoder and \{128, 64, 32, T\} filters in the decoder where $T$ is the number of time steps we predict the CM for.
\vspace{5pt}

\subsection{Training Loss}
The designed architecture is a multi-task network. The objective function is accordingly defined to direct the network to learn each task:
\noindent
\begin{align}
    \mathcal{L}_{total} = w_1 \mathcal{L}_{Pred} + w_2 \mathcal{L}_{CME}
\end{align}

where $w_1$ and $w_2$ are hyperparameters. We define each term in detail below.

\subsubsection{Prediction Loss}

We follow \cite{mohajerin2019multi} to formulate OGM prediction as a classification problem where each cell can be occupied or not. Hence, the objective function includes a pixel-wise Cross-Entropy between the predicted OGMs, $\hat{B^1}$, and the target OGMs, $B^{1*}$, multiplied by a visibility matrix, $\mathcal{V}$, described in \cite{dequaire2018deep} to handle occlusion. Due to unbalanced number of occupied and free cells, we normalize the loss by the ratio of occupied/free cells, $\eta$. Finally, to push the predicted OGMs toward the target OGMs we use Structural Similarity Index Metrics (SSIM) \cite{wang2003multiscale}. The OGM prediction loss is then defined as:

\begin{align}
 \resizebox{1\hsize}{!}{$
 \mathcal{L}_{Pred} = \frac{\eta}{WH} \sum_{x} \sum_{y} \mathcal{V} \odot \mathcal{H}(\hat{B^1}, B^{1*}) + \gamma (1 - SSIM(\hat{B^1}, B^{1^*}))
 $}
\end{align}
\noindent
where $H$ and $W$ are the OGM dimensions, $\odot$ denotes the element-wise product, $\mathcal{H}(a, b)$ is the pixel-wise cross-entropy  and $\gamma$ is a hyperparameter.
\subsubsection{CM Estimation Loss:} Since there is no ground truth for the CM, defining an objective function which pushes the network to learn meaningful CM is challenging. Relying only on the expert's trajectory makes it difficult for the network to generalize. The expert's trajectory only occupies a few cells and it does not give information about the most of the surrounding area. To address these issues we define an objective function consisting of two terms: 

\begin{align}
\mathcal{L}_{CME} = \alpha \mathcal{L}_{p} + \beta \mathcal{L}_{aux}
\end{align}

\noindent
where $\alpha$ and $\beta$ are hyperparameters. 

$\mathcal{L}_{p}$ is defined to inject the prior knowledge about cell cost such as the high cost associated with non-drivable areas. Specifically, $\mathcal{L}_p$ is a classification loss comparing the generated CM and a target ${C}_{target}$, which at each time step is 0 for the cells occupied by the expert and 1 for non-drivable areas and the occupied cells in drivable areas. Since there is no information about the other cells we do not want to push the network to assign any values to them. Moreover, the number of cells belonging to the expert's trajectory (low cost cells) are far less in number than the high cost ones. In order to address both of these issues, we calculate the loss on a subset of cells selected by a mask, $\mathcal{M}$ with 0 and 1 elements. The total number of 1s is set to a predefined number, $N$. $\mathcal{M}$ elements are 1 for the pixels occupied by the expert. The rest of 1s are sampled from the high cost cells, i.e. cells occupied by objects or in non-drivable area, with the high cost cells that are occupied having 2 times more chance to be selected. This ratio empirically speeds up training. The $L_p$ loss function is then:
 
 \begin{align}
 \mathcal{L}_p = \frac{1}{WH} \sum_{x} \sum_{y} \mathcal{M} \odot \mathcal{H}(B_{k}^2, C_{target})
 \end{align}

\noindent

\begin{figure}[t]
	\centering
	\includegraphics[scale=1.1]{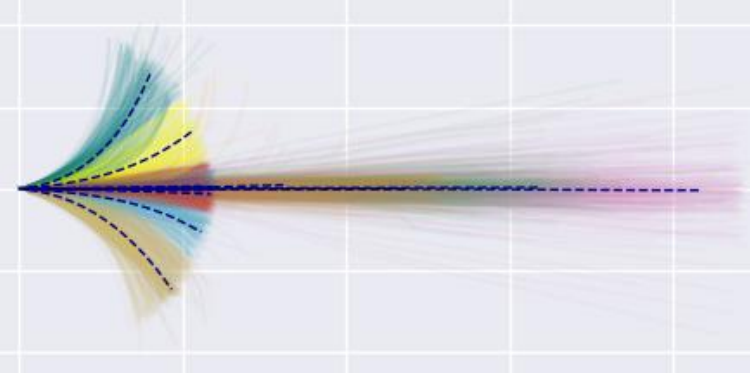}
	\caption{\small Different modes of driving driven from data. Dashed lines show the mean of each cluster.}
	\label{fig:intention}
\end{figure}

We added an \textit{Imitation Network} to the architecture and defined an auxiliary task in overall objective function, $\mathcal{L}_{aux}$, in order to indirectly push the CMs to be a representation of the underlying reason for the expert's behavior. For this purpose, a sequence of estimated CMs from time $\tau$ to the prediction horizon $\tau + T$ are fed to an encoder. These features are then utilized by an \textit{intention prediction} head and a \textit{a regressor head} to predict the expert's trajectory.
 
A predefined set of "intentions" is used to represent different semantic modes made by the expert (e.g. changing lane, speed up, slow down), $\mathcal{I} = \{i^k\}^K_{k=1}$ where $i^k = \{s^k_1, ... s^k_T\}$ defines a trajectory for T timesteps. These intentions are derived by clustering the expert's trajectories in the dataset (Figure \ref{fig:intention}). Specifically, we used the DBSCAN clustering algorithm and Hausdorff distance to cluster trajectories. Given the CMs, for each intention $i^k$, the intention predictor head predicts a Bernoulli distribution $p(i^k|CM_{\tau:T+\tau})$ to determine whether the expert chose that driving mode or not. Hence, each trajectory can belong to multiple clusters at the same time. In this way we do not penalize the network for choosing the modes that are close to each other. One can also use the soft labels in a cross-entropy setting where the labels are the normalized distances to clusters. However, empirically our problem formulation worked better for this architecture. The regressor head then outputs K offsets, $s^o_k$, between the mean of each cluster $\mu_k$ and the expert's trajectory $s^*$. We then used a weighted MSE to optimize the networks. 

\vspace{3pt}
\begin{align} \label{eq:aux}
\mathcal{L}_{aux} = \frac{1}{K}\sum_{k}\mathcal L_{cls}(p_k, p_k^*) + \lambda   \sum_{k}\omega_k MSE(\mu_k + s^o_k ,  s^*)
\end{align}
\vspace{3pt}

\noindent
where ${p}_k$ is the probability of an intention to be the expert's intention, $\omega_k$ is the normalized distance of the groundtruth trajectory to each mode and $\lambda$ is a hyperparameter.

\subsection{Motion Planning} \label{sec:planning}

To evaluate the quality of the CMs, we use them for motion planning. We follow \cite{zeng2019end} to use clothoids \cite{shin1992path} as well as circular and straight lines to define the shape of candidate trajectories. The velocity profile of a candidate trajectory is determined by sampling acceleration in the range of $[-5, 5]$ $m/{s^2}$ and velocity between 0 and the speed limit. Since computing the cost of each candidate trajectory using the estimated CMs is a cheap operation, our motion planning module is computationally very efficient.

Note that the output of the regressor head for the imitation auxiliary task could in principle be used for trajectory planning. However, we opt for a simple sampling method that uses the CMs. This is partly to demonstrate the versatility of the CMs when working with motion planning methods and partly because the CM estimations are presumably much more reliable since they integrate by design broader concerns beyond imitating the expert. 
	\section{Experiments} \label{sec:exp}

\begin{figure*}[!t]
	\centering
	\includegraphics[trim = 0mm 0mm 0mm 0mm,width = 18cm]{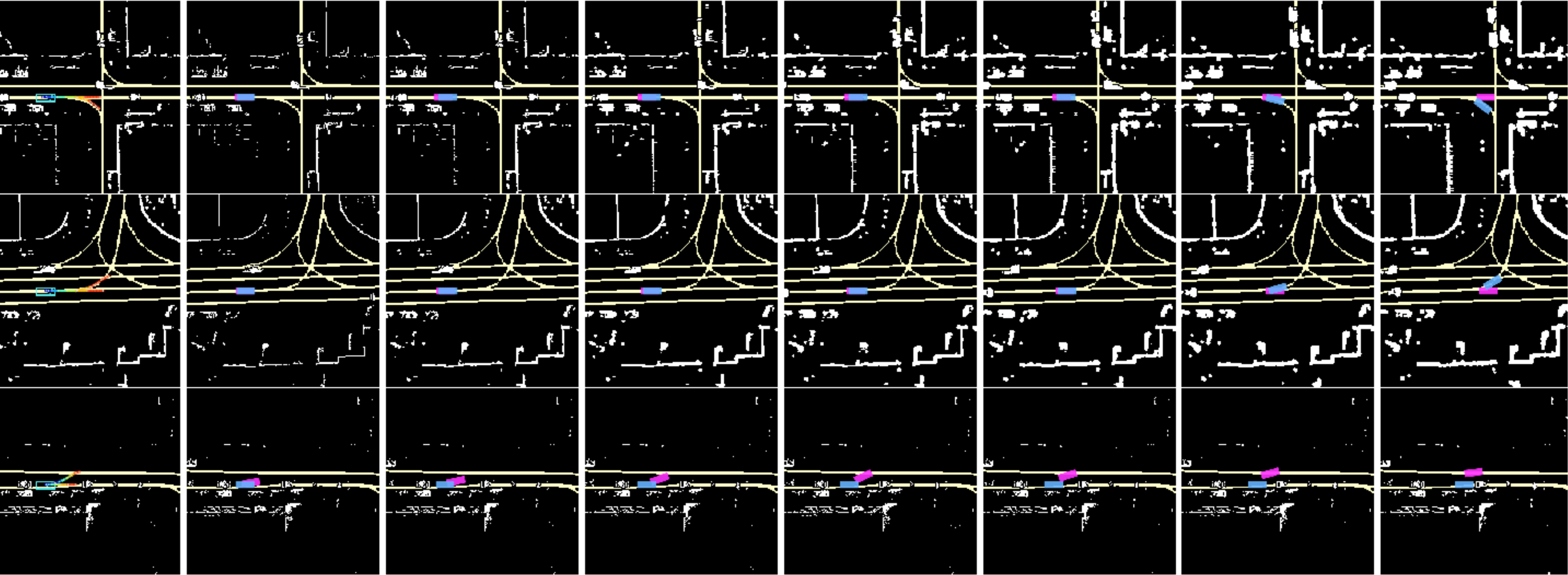}
	\caption{\small Low cost trajectories selected by the proposed method. In these experiments the algorithm is forced to pick the trajectories from different driving modes as described in Section \ref{sec:exp-2-obj}. Each row, is a different scenario where the first column shows the planned trajectory. In next columns the car is moved according to that trajectory.
	}
	\label{fig:results}
\end{figure*}

\begin{table*}
	\footnotesize
	\begin{center}
		\begin{tabular}{|cc||ccc||ccc||ccc|}
			\hline
			\multicolumn{2}{|c||}{Algorithms} & \multicolumn{3}{c||}{$\tau=1$ sec $T=1$ sec} & \multicolumn{3}{c||}{$\tau=2$ sec $T=2$ sec}& \multicolumn{3}{c|}{$\tau=1$ sec $T=3$ sec}\\
			\hline			
			Alg. & Arch. & minADE & CR(\%) & RV(\%)  & minADE & CR(\%) & RV(\%)  & minADE & CR(\%) & RV(\%)\\
			\hline
			
			& \multicolumn{1}{l||}{BC-MLP} & \textbf{0.05} & \textbf{0.00} & 0.07 & \textbf{0.79} & 2.36 & 2.61 & 3.18 & 7.11 & 5.73 \\ 
			\multirow{-2}{*}{BC}& \multicolumn{1}{l||}{BC-LSTM} &  0.08 &  \textbf{0.00} & 0.09 & 0.84 & 2.57 & 1.87 & 3.01  & 6.19 & 5.03\\
			\hline
			
			RuleCM & \multicolumn{1}{c||}{} & 1.03  & 0.34 & \textbf{0.00} & 2.21 & 3.18 & \textbf{0.00} &  3.24 & 4.93 & 0.09 \\
			\hline

			&\multicolumn{1}{l||}{MFP.1}  & 0.21 & \textbf{0.00} & \textbf{0.00}  & 1.92 & 1.18 & 0.86 &  3.78 & 4.33 & 2.98 \\
			\multirow{-2}{*}{MFP$_{K=6}$} & \multicolumn{1}{l||}{MFP.3}  & 0.21 & 0.09 & 0.05 & 1.92 & 2.07 & 0.97 & 3.78 & 5.96 & 3.59 \\
			\hline
			& \multicolumn{1}{l||}{ESP.1}  & 0.41 & 0.21 & \textbf{0.00} & 2.07 & 2.84 & 1.15  & 3.97  & 3.87 & 4.62 \\ 
			\multirow{-2}{*}{ESP$_{K=6}$} & \multicolumn{1}{l||}{ESP.3} & 0.41 & 0.94 & 0.36  & 2.07 & 2.92 & 1.42 & 3.97 & 5.46 & 4.93 \\ 
			\hline\hline
			\rowcolor[gray]{0.9}& \multicolumn{1}{l||}{MSCME.a.1}  & 0.15 & 0.01 & \textbf{0.00}  & 1.94 & 0.92 & 0.01  & 3.52 & 1.68 & 0.05 \\ 
			\rowcolor[gray]{0.9}& \multicolumn{1}{l||}{MSCME.b.1}  & 0.11 & \textbf{0.00} & \textbf{0.00}  & 1.67 & 0.85 & \textbf{0.00} & 3.28 & 1.57 & \textbf{0.01} \\
			\rowcolor[gray]{0.9}& \multicolumn{1}{l||}{MSCME.b.3}  & 0.11 & 0.01 & \textbf{0.00}  & 1.67 & 0.91 & \textbf{0.00} & 3.28 & 1.62 & \textbf{0.01} \\  
			\rowcolor[gray]{0.9}&\multicolumn{1}{l||}{RCME.a.1} & 0.18 & \textbf{0.00} & \textbf{0.00}  & 2.74 & 0.67 & 0.01  & \textbf{2.92} & 0.84 & \textbf{0.01}\\
			\rowcolor[gray]{0.9}& \multicolumn{1}{l||}{RCME.b.1} & 0.17 & \textbf{0.00} & \textbf{0.00}  & 2.81 & \textbf{0.59} & 0.01 & 2.93 & \textbf{0.78} & \textbf{0.01}\\
			\rowcolor[gray]{0.9}\multirow{-6}{*}{CME (ours)}& \multicolumn{1}{l||}{RCME.b.3} & 0.17 & \textbf{0.00} & \textbf{0.00}  & 2.81 & 0.64 & 0.01 & 2.93 & 0.82 & 0.03\\
			
			\hline
		\end{tabular} 

	\end{center}
	\vspace{-5pt}
	\caption{Argoverse dataset planning evaluation. Note that the \textbf{.1} and \textbf{.3} variations using the same models/samples. In \textbf{.1} we selected the trajectory with highest probability/lowest cost. In \textbf{.3} we chose 3. Therefore, the minADE is the same for these variations. Variants of our method (gray) outperformed other algorithm in term of CR in all of the scenarios.}
	\label{tab:Results-}
	\vspace{-10pt}
\end{table*}

\begin{table*}
	\footnotesize
	\begin{center}
		\begin{tabular}{cc||ccc||ccc||ccc}
			\multicolumn{2}{c||}{Algorithms} & \multicolumn{3}{c||}{$\tau=1$ sec $T=1$ sec} & \multicolumn{3}{c||}{$\tau=2$ sec $T=2$ sec}& \multicolumn{3}{c}{$\tau=1$ sec $T=3$ sec}\\
			\hline			
			Alg. & Aux & minADE & CR(\%) & RV(\%)  & minADE & CR(\%) & RV(\%)  & minADE & CR(\%) & RV(\%)\\
			\hline
			\multicolumn{1}{l}{RCME.1}& $\checkmark$ & 0.18 & 0.00 & 0.00  & 2.45 & 0.67 & 0.01  & 2.92 & 0.84 & 0.01\\
			\multicolumn{1}{l}{RCME.1}&  & 0.22 & 0.00 & 0.00 & 2.76 & 0.89 & 0.01  & 3.18 & 4.02 & 0.01\\
			\multicolumn{1}{l}{RCME.3}&$\checkmark$ & 0.18 & 0.00 & 0.00  & 2.45 & 0.93 & 0.01  & 2.92 & 1.00 & 0.01\\
			\multicolumn{1}{l}{RCME.3}& & 0.22 & 0.11 & 0.06  & 2.76 & 2.68 & 0.03  & 3.18 & 7.32 & 0.12\\
			
		\end{tabular}
	\end{center}
	\vspace{-5pt}
	\caption{CME with and without the auxiliary task}
	\label{tab:abl-obj}
	\vspace{-10pt}
\end{table*}

\begin{table*}
	\footnotesize
	\begin{center}
		\begin{tabular}{cc||ccc||ccc||ccc}
			&& \multicolumn{3}{c||}{$\tau=1$ sec $T=1$ sec} & \multicolumn{3}{c||}{$\tau=2$ sec $T=2$ sec}& \multicolumn{3}{c}{$\tau=1$ sec $T=3$ sec}\\
			\cline{3-11}			
			\multicolumn{2}{c||}{\multirow{-2}{*}{CME Methods}} & minADE & CR(\%) & RV(\%)  & minADE & CR(\%) & RV(\%)  & minADE & CR(\%) & RV(\%)\\
			\hline
			\multicolumn{2}{l||}{RCME, with pred} & \textbf{0.15} & \textbf{0.01} & \textbf{0.00}  & \textbf{1.94} & \textbf{0.92} & 0.01  & \textbf{3.52} & \textbf{1.68} & 0.05 \\ 
			\multicolumn{2}{l||}{RCME, without pred}& 0.34 & 0.79 & \textbf{0.00} & 3.13 & 7.18 & \textbf{0.00}  & 3.99 & 11.14 & \textbf{0.00}\\
			
		\end{tabular}
	\end{center}
	\vspace{-5pt}
	\caption{CME with and without the OGM prediction}
	\label{tab:abl-pred}
	\vspace{-10pt}
\end{table*}

\begin{table*}
	\footnotesize
	\begin{center}
		\begin{tabular}{cc||ccc||ccc||ccc}
			&& \multicolumn{3}{c||}{$\tau=1$ sec $T=1$ sec} & \multicolumn{3}{c||}{$\tau=2$ sec $T=2$ sec}& \multicolumn{3}{c}{$\tau=1$ sec $T=3$ sec}\\
			\cline{3-11}			
			\multicolumn{2}{c||}{\multirow{-2}{*}{Algorithms}} & TP & TN & $S_{100}$ & TP & TN & $S_{100}$ & TP & TN & $S_{100}$\\
			\hline
			\multicolumn{2}{c||}{Diff. Learn} &  81.92    & 98.32 &   96.32 & 80.07 & \textbf{99.08} & \textbf{96.52}  & 78.64    & \textbf{99.26}&   \textbf{97.33} \\ 
			\multicolumn{2}{c||}{RCME}& \textbf{82.13}    & \textbf{98.31} & \textbf{97.48}  & \textbf{81.96}  & 97.08 &   93.84  & \textbf{81.38}    &98.19&   95.39\\
			
		\end{tabular}
	\end{center}
	\vspace{-5pt}
	\caption{OGM prediction with and without CME}
	\label{tab:abl-cm}
	\vspace{-10pt}
\end{table*}

We applied our approach to the Argoverse \cite{Argoverse} dataset. The LiDAR point clouds are converted to $256\times256$ BEV with the ground removal described in \cite{miller2006mixture}. Moreover, to increase the ability of our model in handling occlusions, we applied a visibility mask as in \cite{dequaire2018deep}. We also encoded the information from the map into 8 different channels.

We report performance on 3 different settings for the input sequence length $\tau$ and the prediction horizon $T$. In all cases, the data is partitioned into 20 sequences of frames. Thus, the time gap between two consecutive frames for ${(\tau=1, T=1), (\tau=2, T=2), (\tau=1, T=3)}$ are $0.1$, $0.2$ and $0.2$ respectively.

We first evaluate the effectiveness of our approach in trajectory planning using a variety of metrics and compare our approach to multiple baselines. We quantitatively evaluate the performance of all these solutions in different planning horizons. In Section \ref{sec:exp-2} we provide multiple ablation studies to show the effects of different modules and objective terms in the overall performance of the system.

\subsection{Cost Map Estimation} \label{sec:exp-1}

Direct evaluation of predicted CMs is not straightforward as there is no groundtruth for them. We thus evaluate their quality by using them with the planning approach described in Section \ref{sec:planning} and compare planning quality under the following metrics.

\begin{itemize}
	
	\item \textbf{minADE:} We used the minimum average displacement error (minADE) $\min{ \frac{1}{T} \sum_{\tau}^{\tau + T} {\left\lVert{\hat{s} - s^*}\right\rVert}_2}$ to measure the minimum drift of the trajectories generated by each model from the groundtruth. This metric is especially suitable for baselines producing multiple trajectories as well as the proposed method because it does not penalize the trajectories that are valid, but far from the groundtruth. For algorithms that only produce a single trajectory, minADE reduces to ADE. Since our solution aims at capturing the underlying reasons for the expert's behavior rather than merely generating trajectories, just minADE with respect to expert's trajectories is inadequate. 
	
	\item \textbf{Potential Collision Rate (CR):} Each selected trajectory is mapped to the future frames to check if it collides with any object in the scene. While the ego vehicle's behavior affects other cars' trajectories in the real world, for short horizons considered here we may ignore such interaction.

	\item \textbf{Road Violation (RV):} The selected trajectory is mapped to a drivable area to check for possible violations of traffic rules.
	
\end{itemize}
\noindent

We compare our solution with the following baselines: 

\begin{itemize}
	\item \textbf{Behavior Cloning(BC):} We implemented a BC learner that receives a sequence of OGMs and the past trajectory for $\tau$ timesteps and generates trajectories close to those of the human driver. For fair comparison we use the same OGM predictor in our model. Trajectories are generated with one of two architectures, where BC-MLP uses four CNN layers with [32, 64, 64, 128] filters followed by three mlp layers with [64, 32, 2T] units and BC-LSTM uses a CNN encoder with [16, 32] filters to encode map and predicted ($\hat{O}$) or observed ($O^*$) OGMs and predict $S$ for T timesteps.
	
	\item \textbf{Rule-Based Cost Map (RuleCM):} Instead of predicting CMs we use hard rules to shape a CM. We assign high cost to non-drivable areas and the occupied cells at the present time. The same trajectory generator as in Section \ref{sec:planning} is used and driving trajectory with lowest cost is selected. This baseline highlights the importance of the \textit{predicted} cost for motion planning.
	
	\item \textbf{Estimating Social-forecast Probability (ESP):} We compare our results to ESP \cite{rhinehart2019precog} for the single agent, using the code published at \url{https://github.com/nrhine1/precog}. We did not do hyperparameter tuning, but we used both forward KL and symmetric cross entropy for the objective function and reported the best results. We sample 6 trajectories $(K=6)$ for evaluation. For CR and RV we chose top-1 and top-3 trajectories according to the model-assigned probability and reported the results for both. These variations are referred to as ESP.1 and ESP.3. Note that because we use the same samples from the same model to study if all of the generated trajectories are useful for planning, the minADE is the same for these variations. 
	
	\item \textbf{Multiple Future Prediction (MFP):} MFP \cite{tang2019multiple} is a  multi-modal trajectory prediction solution. We use code from \url{https://github.com/apple/ml-multiple-futures-prediction} with adaptations to make it work for a single agent. We acknowledge that this adaptation affects the performance of MFP as the other agents' trajectories are the key inputs to this algorithm. In a complete AD system, such data may come from perception modules that detect and track other agents. Since we are studying the performance of prediction in the absence of such modules, we choose to test MFP in a limit case. We used 3 modes for MFP. Similar to ESP, we use two variations of sampled trajectories, where MFP.1 and MFP.3 refer to top-1 and top-3 trajectory selections respectively.
	
\end{itemize}

For our solution, we study the performance of both the RCME and the MSCME architectures. In one setting (variation \textbf{.a}) we use the trajectories generated according to Section \ref{sec:planning}, and in another setting (variation \textbf{.b}) we add the trajectories generated by the imitation network to the samples. Similar to ESP and MFP, we use top-1 and top-3 trajectories to evaluate the quality of the trajectories.

\subsection{Planning Results}

As shown in Table \ref{tab:Results-}, all solutions have low collision rate (CR) for shorter horizons. As horizon gets longer, history shorter, and frequency lower, CR increases markedly for all the algorithms except for ours.

Both BC baselines have low ADE for shorter horizons $(T=1, 2)$. This is expected as they explicitly minimize the difference between predicted and expert's trajectories. But for the more challenging settings where they plan trajectories for 3 seconds based on 1 second of history, the generated trajectories have higher ADE, CR and RV. In contrast, even though our solution uses a trajectory sampler to propose trajectories it has low ADE, CR and RV in all settings. Adding the predicted trajectory by the imitation network to the samples helps with the performance in some scenarios.

RuleCM has a low RV percentage as we manually assign high values to the non-drivable grids. However, its high CR shows that it cannot handle dynamic objects well. This highlights the importance of the predictive nature of our proposed solution. 

As mentioned above, we did not tune the hyperparameters for MFP and ESP on Argoverse. \cite{park2020diverse} reports better performance for these algorithms but we could not replicate those results. Multiple factors including the difference in hyperparameters, preprocessing of the data, prediction horizon and number of samples may have contributed to this performance gap. We also use the single-agent variant for both, which forces these algorithms to capture interactions using high dimensional inputs only. This can potentially lead to a decline in performance. The high CR of ESP suggests that its architecture may not be effective in capturing dynamics of the environment and interactions from the high dimensional features. Moreover, the increase in CR and RV for top-3 trajectories over top-1 trajectories and also higher CR and RV even when the performance of these algorithms are close to the proposed method in terms of minADE show that the multi-modality of these algorithms may not be directly suitable for planning. In other words, a small minADE is not an indication of the admissibility of all the samples. The authors of ESP used a similar architecture in \cite{rhinehart2018r2p2} and \cite{rhinehart2018deep} for planning; the authors of MFP also did a brief study on using MFP for planning \cite{tang2019multiple}. Given the success of these algorithms in multi-modal trajectory prediction, our experiments suggest that in order to assess their potential in motion planning they should also be evaluated with planning metrics such as CR on real-world data.

\subsection{Ablation Study} \label{sec:exp-2}

\subsubsection{Objective Function}\label{sec:exp-2-obj}

To study the contribution of the proposed auxiliary task we trained two models with or without it and summarized the results from the RCME.a setting in Table \ref{tab:abl-obj}. The results show that for the more challenging scenario ($\tau$ = 3, T = 1) CR is lower when the auxiliary objective is used. Intuitively, the auxiliary task does not affect RV as much because $\mathcal{L}_p$ takes the static environment into account. But as the dynamics of the environment gets more complex, the role of the auxiliary objective gets more clear. 

For the settings with top-3 trajectories, we explicitly chose trajectories from different clusters so as to examine the ability of the system to reason about different scenarios.  This leads to a larger effect of the auxiliary objective (RCME.3 rows in Table \ref{tab:abl-obj}) even in the easier scenarios, suggesting that the auxiliary task contributes to better generalization.

\textbf{Discussion:} We tried to employ the planning objective introduced in \cite{zeng2019end} to compare the results. However, the network failed to estimate CMs. We believe due to the highly sparse nature of that objective, the perception modules in the architecture are crucial to lead the training.
\subsubsection{Network Architecture}
We also conducted ablation studies on different modules in our architecture. 
First, we replace the OGM predictor in the MSCME architecture with a CNN encoder so that the model directly estimates the CM without the help from OGM predictions. We do this only to MSCME, because in RCME the CM estimation is embedded inside the OGM prediction system. The results are summarized in Table \ref{tab:abl-cm}.
This change leads to a large performance gap in terms of CR, suggesting that simultaneous OGM prediction extracts better mid-level features suitable for reasoning about environment dynamics. 

While the quality of OGM prediction is not the focus of this paper, studying it offers more insight into the overall performance of the system. Thus, we compared the OGM predictions of \textbf{RCME} with the \textbf{Difference Learning Architecture} in \cite{mohajerin2019multi}. The metrics we use are percentage of True Positive (TP), True Negative (TN). We also multiply SSIM by 100 ($S_{100}$) to make it the same scale as the other metrics. The results are summarized in Table \ref{tab:abl-pred}. It is not surprising that RCME has higher TP rate compared to Difference Learning, because the additional CM estimation task brings more information to the system, leading to more accurate predictions.
	\section{Conclusion}

 The definition of driving cost is highly ambiguous and should encode human's driving behavior as well as the environment characteristic such as road structure. In this work, we proposed a novel fully self-supervised approach for estimating high dimensional CMs. Due to the importance of the prediction in CME, part of the network is dedicated for predicting high-dimensional OGMs. Input and predicted OGMs, contextual information as well as the human driving behavior are then utilized to extract features required to encode expert's demonstration. We applied the proposed method to Argoverse dataset and illustrated the effectiveness of our approach in different planning horizons.



{\small
\bibliographystyle{ieee_fullname.bst}
\bibliography{cost_map.bib}
}

\end{document}